\documentclass[conference]{IEEEtran}
\IEEEoverridecommandlockouts

\usepackage{cite}
\usepackage{amsmath,amssymb,amsfonts}
\usepackage{algorithmic}
\usepackage{graphicx}
\usepackage{textcomp}
\usepackage{xcolor}
\usepackage{bbm}
\usepackage{graphicx} 
\usepackage{amsmath}
\usepackage[linesnumbered,ruled,vlined]{algorithm2e}
\usepackage{makecell}
\usepackage[colorlinks,linkcolor=blue]{hyperref}

\def\BibTeX{{\rm B\kern-.05em{\sc i\kern-.025em b}\kern-.08em
    T\kern-.1667em\lower.7ex\hbox{E}\kern-.125emX}}
\begin{document}

\title{Mapping at First Sense: A Lightweight Neural Network-Based Indoor Structures Prediction Method for Robot Autonomous Exploration
}


\author{
	\IEEEauthorblockN{Haojia Gao$^{1,2,\dag}$, Haohua Que$^{2,\dag}$, Kunrong Li$^{3}$, Weihao Shan$^{2}$,  Mingkai Liu$^{4}$, Rong Zhao$^{2}$, Lei Mu$^{2,5}$, \\ Xinghua Yang$^{6}$, Qi Wei$^{7}$, Fei Qiao$^{2,*}$}
	\IEEEauthorblockA{$^1$ Department of Fan Gongxiu Honors College, Beijing University of Technology, Beijing 100124, China}
	\IEEEauthorblockA{$^2$ SenseLab, Department of Electronic Engineering, Tsinghua University, Beijing 100084, China}
	\IEEEauthorblockA{$^3$ Department of College of Computer Science, Beijing University of Technology, Beijing 100124, China}
	\IEEEauthorblockA{$^4$ School of Software \& Microelectronics, Peking University, Beijing 102600, China}
	\IEEEauthorblockA{$^5$ College of Computer Science and Artificial Intelligence, Southwest Minzu University, Chengdu 610041, China}
	\IEEEauthorblockA{$^7$ Department of Precision Instrument, Tsinghua University, Beijing 100084, China}
	\IEEEauthorblockA{$^\dag$ These authors contributed equally to this work.}
	\IEEEauthorblockA{$^*$ Corresponding author: qiaofei@tsinghua.edu.cn}
}

\maketitle

\begin{figure*}[htbp]
	\centerline{\includegraphics[width=1\textwidth]{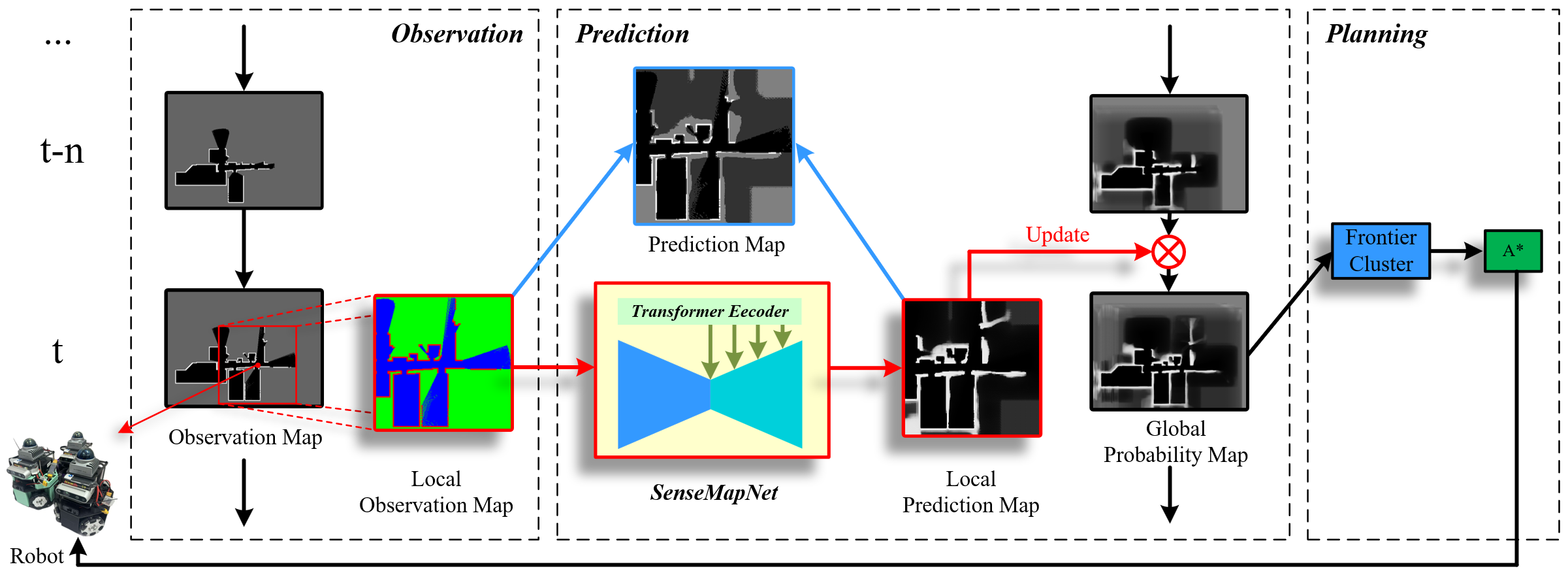}}
	\caption{\textbf{Overview of the SenseMap Pipeline for Autonomous Exploration.}
		This pipeline consists of three main stages: \textit{Observation}, \textit{Prediction}, and \textit{Planning}. At time step $t$, the robot captures an \textit{Observation Map}, extracting a \textit{Local Observation Map} (highlighted in red) from the global representation. The local observation is then processed by the proposed \textit{SenseMapNet}, generating a \textit{Local Prediction Map}, which is further integrated into the \textit{Global Prediction Map}. The updated map is used to identify unexplored frontiers via \textit{Frontier Clustering}, followed by path planning using the \textit{A*} algorithm. This pipeline enhances exploration efficiency by reducing uncertainty and improving map completeness.}
	\label{fig1}
\end{figure*}

\begin{abstract}
	Autonomous exploration in unknown environments is a critical challenge in robotics, particularly for applications such as indoor navigation, search and rescue, and service robotics. Traditional exploration strategies, such as frontier-based methods, often struggle to efficiently utilize prior knowledge of structural regularities in indoor spaces. To address this limitation, we propose \emph{Mapping at First Sense}, a lightweight neural network-based approach that predicts unobserved areas in local maps, thereby enhancing exploration efficiency. The core of our method, \emph{SenseMapNet}, integrates convolutional and transformer-based architectures to infer occluded regions while maintaining computational efficiency for real-time deployment on resource-constrained robots. Additionally, we introduce \emph{SenseMapDataset}, a curated dataset constructed from KTH and HouseExpo environments, which facilitates training and evaluation of neural models for indoor exploration. Experimental results demonstrate that \emph{SenseMapNet} achieves an SSIM (structural similarity) of 0.78, LPIPS (perceptual quality) of 0.68, and an FID (feature distribution alignment) of 239.79, outperforming conventional methods in map reconstruction quality. Compared to traditional frontier-based exploration, our method reduces exploration time by 46.5\% (from 2335.56s to 1248.68s) while maintaining a high coverage rate (88\%) and achieving a reconstruction accuracy of 88\%. The proposed method represents a promising step toward efficient, learning-driven robotic exploration in structured environments.
\end{abstract}


\section{Introduction}  \label{sec 1}
Autonomous exploration is pivotal problem in robotics realm, with significant application in areas ranging from indoor navigation \cite{gonzalez2002navigation} to planetary exploration \cite{vogele2024robotics}. It involves the process of creating environment models through a recursive relationship between sensing and movement. This process introduces several complexities, including the need for real-time decision-making, environment variability, and the challenge of building reliable models of the environment. Efficient exploration strategies typically focus on selecting the most informative areas to explore, which involves estimating about unexplored regions while considering the influence from sensors and environment factors \cite{julia2012comparison}.

The number of real-world environments targeted for exploration exhibit some level of structure, predictability and repetitiveness in their geometric layout \cite{uyttendaele2004image}. Particularly, the indoor environment match with these factors, such as offices building or hospitals, are often composed of repeated rooms and corridors. By observing partial information from one area of the environment, it is possible to predict overall information of this area or even other areas. This inherent predictability can enhance exploration efficiency, allowing robots to leverage partial observation to guide navigation and mapping efforts.

While traditional exploration methods have made significant strides, many still struggle to fully leverage environment predictability in real-time setting \cite{ingrand2017deliberation,chen2024survey}. Conventional methods, such as frontier-based exploration \cite{yamauchi1998frontier}, typically using a greedy strategy. In environment with lower complexity, this method is generally sufficient to address most exploration tasks. However, in more information-rich environments, these frontiers fail to effectively guide exploration.

To address these limitations, recent advancements have incorporated machine learning techniques to improve exploration. Many of these methods explicitly predict map uncertainty and navigate towards these uncertain regions \cite{georgakis2022uncertainty}. While this approach leverages uncertainty in the map predictions, it does not take into account errors introduced by the sensor's field of view (FOV). Additionally, some methods employ large models to predict the overall environment \cite{ho2024mapex}, but these approaches often overlook the computational constraints of the robot's onboard hardware, potentially leading to significant computational burdens that can reduce efficiency.

In this paper, We propose \emph{Mapping at First Sense} (It will be referred to simply as \emph{SenseMap} hereafter in this paper), a novel approach that integrals a lightweight neural network for real-time local map prediction of unobserved areas. By focusing on reducing the computational burden and the performance of autonomous exploration, our method offers a promising solution for autonomous robotics systems.

Our contributions are summarized as follows:
\begin{itemize}
	\item We introduce a lightweight neural network model (\emph{SenseMapNet}) for efficiently predicting local map in real-time.

	\item The development of a novel method that combines prediction of unobserved areas with real-time exploration strategies.

	\item We present an indoor dataset (\href{https://github.com/senselabrobo/SenseMapDataset}{\emph{SenseMapDataset}}) based on KTH dataset \cite{aydemir2012can} and HouseExpo platform \cite{li2020houseexpo}, to demonstrate the effectiveness of our proposed approach.
\end{itemize}

This paper is organized as follows. Section~\ref{sec 1} and Section~\ref{sec 2} introduce the background and related works, respectively. Section~\ref{sec 3} provides a problem definition of the proposed system. \emph{SenseMapNet} is described in detail in Section~\ref{sec 4}. Experimental results and discussion are presented in Section~\ref{sec 5}. Finally, a conclusion is drawn in Section~\ref{sec 6}.

\section{Related Work} \label{sec 2}
\subsection{Autonomous Exploration} \label{sec 2.2}
Autonomous exploration in unknown environments has been extensively studied using traditional methods, particularly the frontier-based approach. The seminal work by Yamauchi introduced the frontier-based method, which is widely used for autonomous robot exploration \cite{yamauchi1998frontier}. This method identifies the boundary between explored and unexplored areas (the frontier) and guides the robot to move towards these frontier points. One of the simplest implementations of this method involves using the breadth-first search (BFS) algorithm \cite{bundy1984breadth} to find paths to each frontier and selecting the center of the frontier with the shortest path as the target point \cite{hing2015breadth,indriyono2021optimization,rahim2018breadth}. However, this method may lead to collisions if there are obstacles near the selected target point. To mitigate this, González-Baños et al. proposed moving the target point slightly into the free space to avoid collisions \cite{gonzalez2002real}.

Juliá et al. \cite{julia2012comparison} pointed out that the autonomous exploration problem of robots in unknown environments is a partial observation Markov decision process (POMDP) . This perspective has led to the development of various strategies that aim to optimize the exploration process by considering factors such as information gain and path length. Traditional methods often use one or more objective functions to choose which frontier to explore next, balancing the trade-off between exploration efficiency and computational complexity.

Coordination among multiple robots has also been explored to enhance exploration efficiency. Burgard et al. proposed centralized systems where a single entity coordinates the actions of multiple robots, offering comprehensive control but suffering from scalability and single-point-of-failure issues \cite{burgard2005coordinated}. In contrast, Yamauchi demonstrated distributed systems where robots make autonomous decisions based on local information, which are more scalable and resilient but face communication constraints and coordination complexities \cite{yamauchi2020distributed}. Cooperative strategies such as environmental segmentation, auction-based assignment, and Sequential Greedy Assignment (SGA) have been developed to optimize resource allocation and coordination in multi-robot exploration \cite{ribeiro2024efficient,athira2024systematic}.

\subsection{Neural Network-Based Navigation} \label{sec 2.1}
Recent advancements in neural network-based navigation have significantly improved the capabilities of autonomous systems across a wide range of applications \cite{katkuri2024autonomous}. By harnessing the power of neural networks, these methods enhance navigation accuracy, adaptability, and efficiency, particularly in complex and dynamic environments.

One significant avenue of research involves leveraging neural networks for path planning. Due to their high robustness, neural networks can outperform traditional methods in more complex environments by generating optimized paths. By feeding an obstacle map into a neural network, the system can predict a feasible exploration path, enhancing efficiency. Qureshi et al. introduced Motion Planning Networks, a framework that enables robots to rapidly generate feasible trajectories even in intricate environments \cite{qureshi2019motion}. Vidal et al. extended the application of neural network-based path planning to Autonomous Underwater Vehicles (AUVs), demonstrating its efficacy in underwater navigation \cite{vidal2019online}. Lee et al. encoded indoor environments using an ENN network and combined deep Q-learning with convolutional neural networks (CNNs) to achieve efficient motion planning, allowing robots to explore indoor spaces with high adaptability \cite{lee2021extendable}. Cimurs et al. further advanced this direction by integrating deep reinforcement learning with global navigation strategies, making the system more adept at handling both complex and dynamic environments \cite{cimurs2021goal}.

Another key research direction focuses on leveraging neural networks for perception, prediction, and reconstruction, enabling the generation of obstacle maps that can then be used for subsequent path planning. Instead of relying on traditional map-building techniques, this approach directly processes sensor data through neural networks to infer obstacle distribution. Shrestha et al. employed convolution and transposed convolution to predict spatial boundaries, using the resulting maps to compute information gain, thereby optimizing the exploration process \cite{shrestha2019learned}. Wang et al. took a step further by training a neural network to output information maps that not only represent the environment but also encode localization uncertainty, offering a more comprehensive understanding of the robot's surroundings \cite{wang2022information}. Meanwhile, Ho et al. \cite{ho2024mapex} maintained multiple instances of the LaMa model \cite{suvorov2022resolution} to predict global maps, subsequently computing information gain to guide navigation.

Recent research has also explored the integration of neural networks into simultaneous localization and mapping (SLAM) systems for autonomous ground vehicles \cite{saleem2023neural}. A comprehensive review of these advancements highlights how deep learning techniques can effectively overcome the limitations of conventional visual SLAM, particularly in scenarios demanding high accuracy and real-time performance.

While these studies have made significant contributions to enhancing exploration efficiency in complex environments, they often overlook the computational overhead associated with neural network inference. This limitation becomes particularly relevant in edge-computing scenarios, where onboard computational resources are constrained. Addressing this gap, the present study focuses on robotic exploration and reconstruction in environments with limited computational capacity. The objective is to strike a balance between maintaining exploration efficiency and reducing computational demands, ensuring that neural network-based solutions remain practical for real-world deployment in resource-constrained settings.

\section{Problem Definition} \label{sec 3} 
To address the problem of robot exploration in unknown indoor environments, we conducted a series of assumptions for better analysis. We assume that the indoor environment \( E \) is a two-dimensional space, and the true map of the environment is unknown, represented by \( M_{true} \in \mathbb{R}^2 \). A two-dimensional map \( M \) is constructed in this two-dimensional space, using coordinates \( a,b \) to describe the value at time \( t \) in the map \( M^t(a,b) \), where the values are:

\begin{equation}  \label{eq1}
	M^t(a, b) = \begin{cases}
		0,   & \text{if the point is free},        \\
		0.5, & \text{if the point is uncertain},   \\
		1,   & \text{if the point is an obstacle}.
	\end{cases}
\end{equation}

At time \( t \), the robot's position is \( P^t = (x, y) \), and the robot's state \( S^t(P^t, \gamma^t) \) is described by the robot's position \( P^t \) and the robot's orientation \( \gamma^t \in (-\pi, \pi] \).

The robot starts from a location \( P^t \), equipped with a 360° LiDAR sensor with range \( L \). At the initial state \( t_0 \), we set \( \forall M^{t_0} (a,b) = 0.5, a,b \in \mathbb{R} \).

When the robot's state is \( S^t(P^t) \), the LiDAR sensor emits beams from the robot's position \( P^t \), scanning at 360°. When the LiDAR beam does not encounter obstacles, the set of areas affected by the LiDAR beam at a distance $L$ is denoted as $A$, the corresponding region in the map is $M^t(a, b) \leftarrow 0$, for $a, b \in \mathbb{A}$. If the LiDAR beam is blocked by an obstacle at the point $O(a_o, b_o)$ within a distance $L$, the obstacle is marked on the map as $M^t(a_o, b_o) \leftarrow 1$. The remaining points covered before being blocked are denoted as $B$, where $M^t(a, b) \leftarrow 0$, for $a, b \in \mathbb{B}$.

The objective of this problem is to find the shortest path $s$ and the shortest time $t$ under the given environment $E$, the robot's initial state $S^{t_0}$, and the LiDAR range $L$, such that the error between the constructed global map $M$ and the ground truth map $M_{{true}}$ is minimized.
\begin{figure*}[htbp]
	\centerline{\includegraphics[width=1\textwidth]{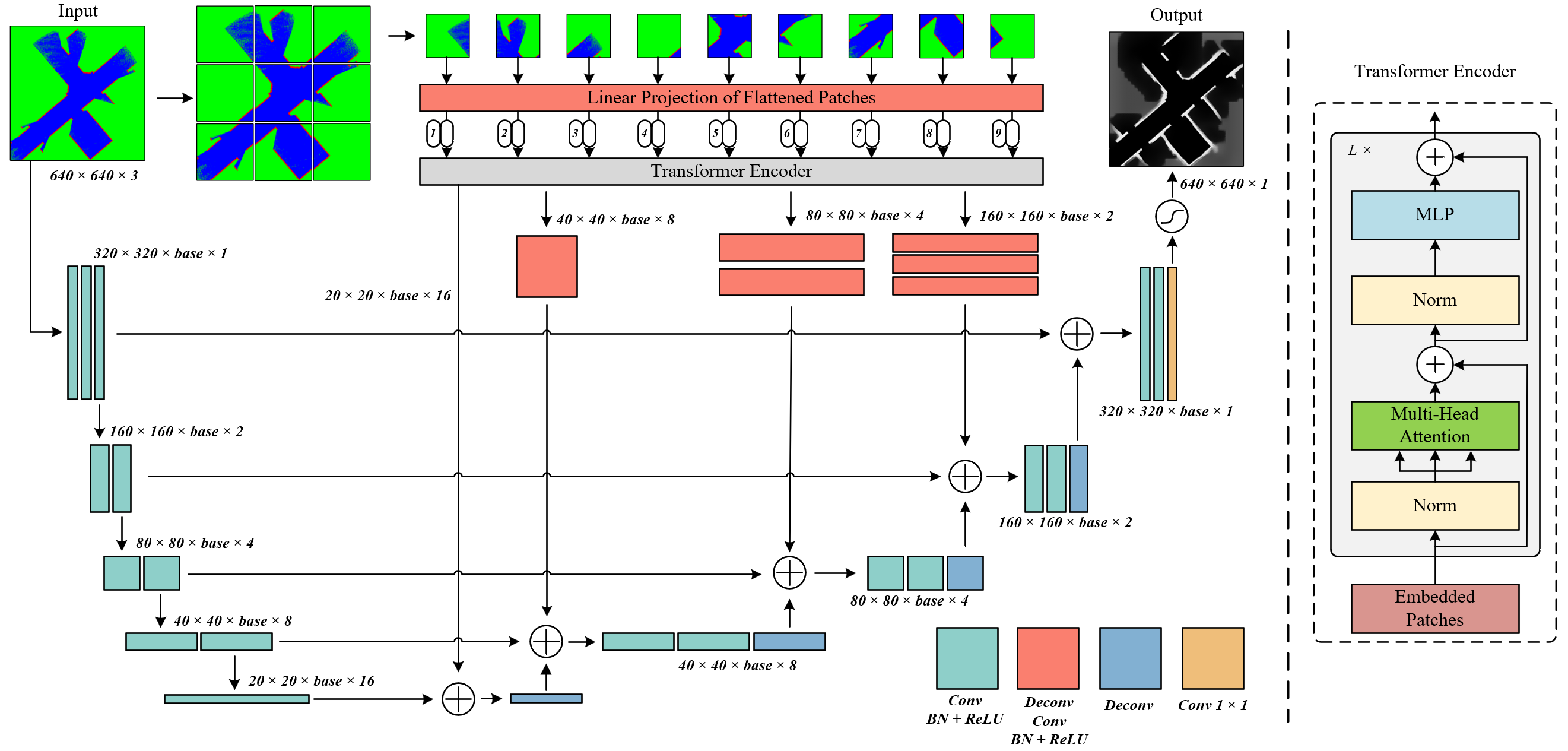}}
	\caption{\textbf{Architecture of SenseMapNet for Local Map Prediction.}
		The input local observation map undergoes two parallel processing streams: a convolutional encoder-decoder network and a Transformer-based encoding pipeline. The convolutional encoder extracts hierarchical spatial features, progressively reducing the resolution while increasing feature depth. Simultaneously, the input map is divided into non-overlapping patches, which are flattened and projected into an embedding space before being processed by the Transformer Encoder. The Transformer module captures long-range dependencies and global spatial relationships. The extracted multiscale features from both streams are fused through skip connections and multi-resolution aggregation, followed by a decoding process to reconstruct the local prediction map. This dual-branch structure enables the model to leverage both fine-grained local spatial details and high-level contextual information, improving prediction accuracy and robustness in autonomous navigation tasks.}
	\label{fig2}
\end{figure*}

\section{Approach} \label{sec 4}
In order to achieve efficient autonomous exploration in indoor environments, we designed an innovative approach, and pipeline as shown in Fig.~\ref{fig1}. Our goal is to use neural networks to accurately predict areas where the current LiDAR sensor data fails to cover.

We input the observed local map in the form of images to the neural network, and the probability of obstacle (In this study, the obstacles are mostly walls) is output the through the network. In Section~\ref{sec 4.1}, we introduce the \emph{SenseMapDataset} and explain how the observed local map serves as input to the neural network. Section~\ref{sec 4.2} provides an introduction to our designed neural network, \emph{SenseMapNet}, and Section~\ref{sec 4.3} details how map construction is carried out based on the output from the neural network.
\subsection{Local Observation Map Processing} \label{sec 4.1}

We define the local observation map \( M_{obs}^t \) at time \( t \), centered around the robot's position \( P^t \), with a sensing range of \( L \). The local map is constructed in the entire map \( M \) with the observation area defined by twice the sensing range \( L \), resulting in the local observation map:
\begin{equation} \label{eq2}
	M^{t}_{{obs}} = M(a, b \mid a \in [P^{t}_x - L, P^{t}_x + L], b \in [P^{t}_y - L, P^{t}_y + L])
\end{equation}

LiDAR provides accurate observations within its range, but it is limited by its coverage and occlusions caused by obstacles. Therefore, by setting the local map with a side length of $2L$, we can provide the model with a certain prediction space without significantly increasing the input size.

After obtaining the local observation map \( M_{obs}^t \), we will assign the codes for free, uncertain, and obstacle areas, mapping them to the corresponding model features. These inputs are used for the model training. Additionally, during dataset construction, we applied the cropping method described in Eq.~\ref{eq2} to extract the corresponding ground-truth labels for each local observation map.

We modified the simulation environment proposed by HouseExpo \cite{li2020houseexpo} to enable the collection of local observation maps and ground truth labels. Using the modified simulation environment, we performed simulations on the KTH dataset \cite{aydemir2012can}, creating the local map prediction dataset, \emph{SenseMapDataset}. This dataset will be used for subsequent training and comparative experiments with our network model, which will be introduced in Section~\ref{sec 5.1}.

\subsection{Local Map prediction Model} \label{sec 4.2}

As the model will be deployed on resource-constrained robotic devices, efficient neural networks are essential.
Building upon the UNet \cite{ronneberger2015u} + Transformer \cite{vaswani2017attention} framework, we incorporated the principles of Vision Transformer (ViT) \cite{dosovitskiy2020image} to seamlessly integrate Transformer-based designs \cite{vaswani2017attention} into image processing. Drawing insights from existing UNet + Transformer architectures \cite{hatamizadeh2022unetr,kaviani2023image,chen2021transunet,petit2021u} we refined and designed \emph{SenseMapNet}, an architecture tailored specifically for local map prediction. The overall structure of \emph{SenseMapNet} is depicted in Fig.~\ref{fig2}.

The UNet \cite{ronneberger2015u} network demonstrates strong capabilities in local feature extraction. Through skip connections, it is able to recover spatial information. The Transformer \cite{vaswani2017attention} model, on the other hand, excels at contextual understanding and global perception. By combining the output of the Transformer Encoder with the skip connections, the map prediction performance can be further enhanced.

The network takes $3 \times H \times W$ map data as input, where the three channels correspond to free, uncertain, and obstacle areas. After inputting the map into the model, the data is processed in patches.
These patches undergo linear projection of flattened patches, which are then fed into the Transformer Encoder as flattened patches and into the convolutional Encoder while retaining their original image format. The output from the Transformer Encoder is divided into four equal parts, which are passed through different layers of Deconvolution to upscale them to different shapes. These are then concatenated with the skip connections from the convolutional Encoder, providing global perceptual information for convolutional layers at different scales.
Through several layers of convolutional decoders, the output is an $H\times W \times 1$ obstacle prediction map, which enables the prediction of the local map.

The model can adjust the input and output channels of the convolutional layers and the embedding dimension of the Transformer Encoder by modifying the base value. This flexibility allows for easy adjustment of the model size. In subsequent experiments, we will compare the performance of the standard model with a base value of 16 and the large model with a base value of 32.

\subsection{Loss function} \label{sec 4.3}

The loss function should better reflect the model's ability to predict unknown areas in the local observation map. Traditional regression loss functions aim for perfect pixel-wise matching between the model and the ground truth. However, in the context of local map prediction tasks, the free area typically occupies a larger proportion of the label map compared to the obstacle area. Pursuing perfect pixel-wise matching with the label can lead to ``conservative'' model that overly resembles the observation map, limiting its ability to effectively predict the unknown regions.

\begin{equation} \label{eq3}
	\ell_{{mse}}(\hat{m}, m) = \frac{1}{N} \| \hat{m} - m \|_2^2
\end{equation}

To address this issue and enhance the model's prediction ability in unknown environments, we leverage Feature Reconstruction Loss \cite{johnson2016perceptual} for training. Feature Reconstruction Loss uses a trained loss network $\varphi$ to evaluate the feature reconstruction error between predicted map images and target images.
The Feature Reconstruction Loss evaluates the feature reconstruction loss between the predicted and target images using a pre-trained loss network $\varphi$. This encourages the model's predictions to exhibit similar structural features as those calculated by the loss network $\varphi$. The loss network $\varphi$ is a deep convolutional neural network with $L$ layers. Let $\varphi_j(m)$ denote the activations at the $j$-th layer of the loss network $\varphi$ when processing the map $m$; if $j$ is a convolutional layer, $\varphi_j(m)$ will be a feature map of shape $C_j \times H_j \times W_j$. The feature reconstruction loss is defined as the Euclidean distance between the feature representations.

\begin{equation} \label{eq4}
	\ell_{{feat}}^{\varphi}(\hat{m}, m) = \frac{1}{L} \sum_{j=1}^{L} \frac{1}{C_j \times H_j \times W_j} \| \varphi_j(\hat{m}) - \varphi_j(m) \|_2^2
\end{equation}

The Mean Squared Error (MSE) loss and the Feature Reconstruction Loss are each multiplied by their corresponding weights, $w_{{mse}}$ and $w_{{feat}}$, and then summed together to obtain the final loss $l(\hat{m}, m)$.

\begin{equation} \label{eq5}
	\ell(\hat{m}, m) = w_{{mse}} \times \ell_{{mse}}(\hat{m}, m) + w_{{feat}} \times \ell_{{feat}}^{\varphi}(\hat{m}, m)
\end{equation}

In our experiments, the loss network $\varphi$ is a 16-layer VGG \cite{simonyan2014very} network pre-trained on ImageNet \cite{russakovsky2015imagenet}, with $w_{{mse}} = 10$ and $w_{{feat}} = 1$. Subsequent experiments revealed that the Feature Reconstruction Loss played a crucial role in the local map reconstruction task.

\subsection{Exploration based on Predicted Map} \label{sec 4.4}

Traditional map exploration algorithms typically rely on maps of known and unknown areas. In contrast, \emph{SenseMap} leverages the model's ability to predict maps, maintaining a global probability map $M_{{PROB}}^t$, thereby transforming the exploration algorithm to rely on the global probability map.

We initialize the global probability map $M_{{PROB}}^0$ with a value of 0.5. At time $t$, the map is updated based on the robot's position $P^t$, using the local predicted map $M_{{pred}}^t$ and the update weight $\alpha$. Let the side length of $M_{{pred}}^t$ be $\lambda$, then:
\begin{equation}  \label{eq6}
	\begin{IEEEeqnarraybox}[][c]{rCl}
		M_{{PROB}}^t (a_i^t, b_j^t) &=& \alpha \times M_{{pred}}^t (i, j) + (1 - \alpha) \times M_{{PROB}}^{t-1} (a_i^t, b_j^t), \\
		&& {s.t. } \quad
		\begin{cases}
			a_i^t = P_x^t - \lambda + i, \\
			b_j^t = P_y^t - \lambda + j, \\
			i, j \in [0, \lambda].
		\end{cases}
	\end{IEEEeqnarraybox}
\end{equation}

After obtaining the global probability map $M_{{PROB}}^t$, we implement autonomous robot exploration using a modified version of the Frontier-Based algorithm proposed by Yamauchi et al \cite{yamauchi1998frontier}.

\begin{algorithm}[htbp]
	\caption{SenseMap: Target Acquisition}
	\label{algorithm1}
	\KwIn{$M_{{PROB}}^t$, $S^t$}
	Extract frontier points set $\mathbb{F} \leftarrow M_{{PROB}}^t$ \\
	\If{$\mathbb{F}$ is not empty}{
	Cluster frontier points: $\mathbb{C} \leftarrow \mathbb{F}$ \\
	\For{each cluster $C$ in $\mathbb{C}$}{
		Compute cost: $C.{cost} \leftarrow M_{{PROB}}^t$ \\
	}
	Obtain waypoint: $G^i \leftarrow \min(C.{cost}).{center}$ \quad
	}
	\Else{
		Mapping process complete
	}
	\KwOut{$G^i$, Map status (whether complete)}
\end{algorithm}

\begin{algorithm}[htbp]
	\caption{SenseMap: Map Reconstruction}
	\label{algorithm2}
	\KwIn{Initial state $S^0$, maximum time $T$}
	Initialize global probabilistic map $M^0_{PROB}$ \\
	\For{each $t \in T$}{
	\While{$G^i$ is available and $P^{t-1} \neq G^i$}{
	Compute movement path: $Path^t \leftarrow M^{t-1}, S^{t-1}, G$ \\
	Update robot state: $S^t \leftarrow Path^t, S^{t-1}$ \\
	Update map: $M^t \leftarrow S^t$ \\
	Extract local observation map: $M^t_{obs} \leftarrow M^t, S^t$ \\
	Predict using SenseMapNet: $M^t_{pred} \leftarrow M^t_{obs}$ \\
	Update global probabilistic map: $M_{PROB}^t \leftarrow M^t_{pred}, S^t$ \\
	}
	Obtain next waypoint: $G^i \leftarrow M^t_{PROB}, S^t$ \\
	\If{$G^i$ is None}{
		\textbf{break} \\
	}
	}
	Binarize the global probabilistic map to obtain the final map: $M \leftarrow M^{last}_{PROB}$ \\
	\KwOut{Final reconstructed map $M$}
\end{algorithm}

As shown in Alg.~\ref{algorithm1}, we classify all probabilities in $M_{{PROB}}^t$ below the threshold $\tau$ as free, all probabilities above the threshold $\nu$ as obstacles, and those between $\tau$ and $\nu$ as uncertain. Using a breadth-first search (BFS) algorithm, we identify the boundary points of free and uncertain regions. If no boundary points are found, the exploration is considered complete. If boundary points exist, we use a connectivity algorithm to cluster all boundary points and calculate the centroid distance from the robot, denoted as $C.{dis}$. The cost of each cluster is determined by combining the probability values, distances, and their corresponding weights, $w_{{PROB}}$ and $w_{{dis}}$, as shown in Eq.~\ref{eq7}. The waypoint $G^i$ is set to the centroid of the cluster with the minimum cost, and the values of $w_{{prob}}$ and $w_{{dis}}$ are determined based on the map resolution.

\begin{equation} \label{eq7}
	C.{cost} = w_{{PROB}} \times \left| 0.5 - M_{{PROB}}^t(a, b) \right| + w_{{dis}} \times C.{dis}
\end{equation}

In our approach, the update weight $\alpha$ is set to 0.25. Given that the number of free pixels in the training data significantly exceeds that of obstacle pixels, the model exhibits a tendency to predict unknown regions as free space. To mitigate this bias and enhance prediction accuracy, we set the threshold values as $\tau = 0.1$ and $\upsilon = 0.5$. After obtaining the target point $G^i$, the A* algorithm is used to acquire the path $Path^t$ and update the robot's state $S^t$. The aforementioned steps are repeated, as shown in Alg.~\ref{algorithm2}, until the entire map is reconstructed.

\section{Experiment} \label{sec 5}
We conduct comparison experiments on the loss function and model network using \emph{SenseMapDataset}. In the loss function comparison experiments, the results were improved by introducing perceptual loss. In the comparison experiments for local map prediction tasks using different neural network models, \emph{SenseMapNet} performed better. Finally, we compared the \emph{SenseMap} method with other exploration methods, which showed satisfactory results.

\subsection{Data Preparation and Experiment Setting} \label{sec 5.1}
\begin{figure}[htbp]
	\centerline{\includegraphics[width=0.4\textwidth]{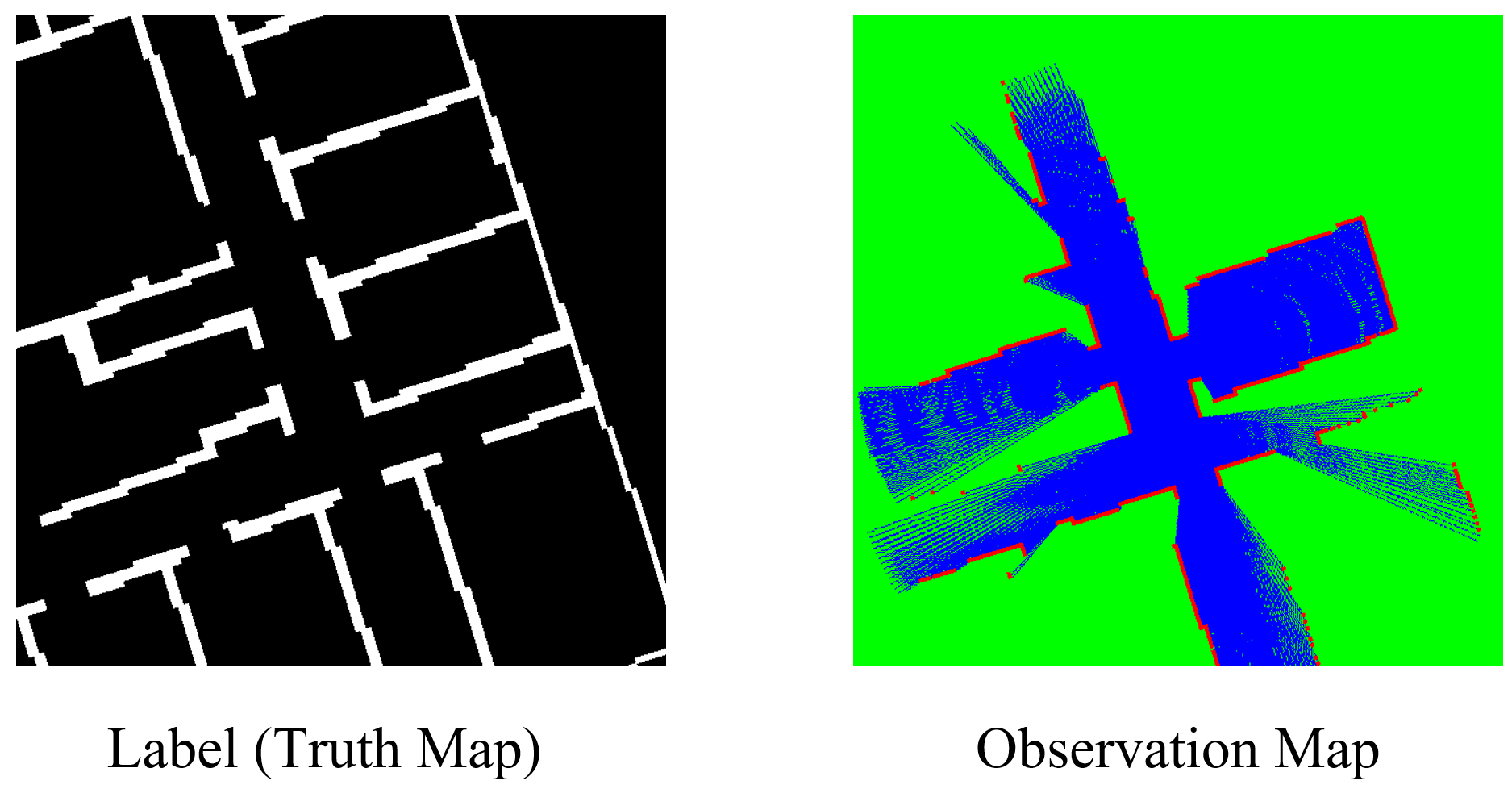}}
	\caption{\textbf{Example of dataset samples.}
		The left image represents the ground truth label map, where white pixels indicate free space and black pixels denote obstacles. The right image illustrates the corresponding observation map, obtained from the robot's sensor data. The observation map consists of three color-coded channels: blue for free space, green for uncertain regions, and black for unexplored areas. This dataset provides rich spatial information, enabling the model to learn and predict local map structures effectively.}
	\label{fig3}
\end{figure}

This study introduces a series of methodological enhancements to the simulation framework adapted from HouseExpo \cite{li2020houseexpo}, with key modifications addressing both functional and algorithmic dimensions. Specifically, we have developed a two-tier mapping system comprising real-time partial observation maps and their corresponding ground truth counterparts, establishing a dual-layer spatial representation framework. Building upon this infrastructure, the classical A* pathfinding algorithm has been strategically integrated to enable optimal trajectory planning within dynamically constrained environments.

The \emph{SenseMapDataset} contains 23,049 data samples, with all maps saved in PNG format. The local observation maps are $640 \times 640$ pixel color maps with 3 channels, where the pixel values are either 255 or 0. A pixel with a blue channel value of 255 represents free space; a pixel with a green channel value of 255 represents uncertain space; and a pixel with a red channel value of 255 represents an obstacle. The local ground truth maps are $640 \times 640$ pixel single-channel grayscale maps, where only the values 255 and 0 are present. Here, 255 represents an obstacle, and 0 represents free space. Fig.~\ref{fig3} shows sample data from the dataset.

\begin{table}[htbp]
	\caption{Distribution of the three channels in the local observation map (in pixels)} \label{tab1}
	\begin{center}
		\resizebox{\columnwidth}{!}{
			\begin{tabular}{|c|c|c|c|c|c|}
				\hline
				\textbf{Category} & \textbf{Mean} & \textbf{Max} & \textbf{Min} & \textbf{Std} & \textbf{Var} \\
				\hline
				Free              & 107041.81     & 226004.00    & 41082.00     & 28878.11     & 833945433.75 \\
				\hline
				Uncertain         & 296115.86     & 363392.00    & 173873.00    & 29356.15     & 861784000.00 \\
				\hline
				Obstacle          & 6442.32       & 11504.00     & 5120.00      & 1046.59      & 1095351.46   \\
				\hline
			\end{tabular}
		}
	\end{center}
\end{table}

\begin{table}[htbp]
	\caption{Distribution of the feature values in the ground truth map (in pixels)} \label{tab2}
	\begin{center}
		\resizebox{\columnwidth}{!}{
			\begin{tabular}{|c|c|c|c|c|c|}
				\hline
				\textbf{Category}    & \textbf{Mean} & \textbf{Max} & \textbf{Min} & \textbf{Std} & \textbf{Var}  \\
				\hline
				Free                 & 317179.25     & 401229.00    & 119897.00    & 51082.56     & 2609427612.08 \\
				\hline
				Obstacle             & 92420.75      & 289703       & 8371.00      & 51082.56     & 2609427612.08 \\
				\hline
				Radio(free/obstacle) & 4.97          & 47.93        & 0.41         & 3.46         & 11.98         \\
				\hline
			\end{tabular}
		}
	\end{center}
\end{table}

\begin{table}[htbp]
	\caption{Coverage of the local observation map on the local ground truth map (coverage of the free pixels in the local observation map/ground truth map)} \label{tab3}
	\begin{center}
		\begin{tabular}{|c|c|c|c|c|c|}
			\hline
			\textbf{Category} & \textbf{Mean} & \textbf{Max} & \textbf{Min} & \textbf{Std} & \textbf{Var} \\
			\hline
			Coverage          & 0.34          & 0.94         & 0.11         & 0.10         & 0.01         \\
			\hline
		\end{tabular}
	\end{center}
\end{table}

Tab.~\ref{tab1} shows the distribution of pixel values in the three channels of the local observation map. Tab.~\ref{tab2} shows the distribution of obstacle and free pixels in the local ground truth map. Tab.~\ref{tab3} shows the coverage of the local observation map on the local ground truth map.

In our experiment, the experimental platform was an AMD EPYC 7502P CPU, NVIDIA RTX TITAN 24GB GPU * 4, and 128GB DDR4 RAM. We used 17,252 samples from the \emph{SenseMapDataset} as the training set and the remaining 5,797 samples as the testing set for the comparative experiment. The Adam optimizer was used, and the training lasted for a total of 300 epochs. For the first 150 epochs, the learning rate was set to 0.001, and for the remaining 150 epochs, the learning rate was reduced to 0.0001.

\subsection{Comparison of Loss function} \label{sec 5.2}

\begin{figure}[htbp]
	\centerline{\includegraphics[width=0.5\textwidth]{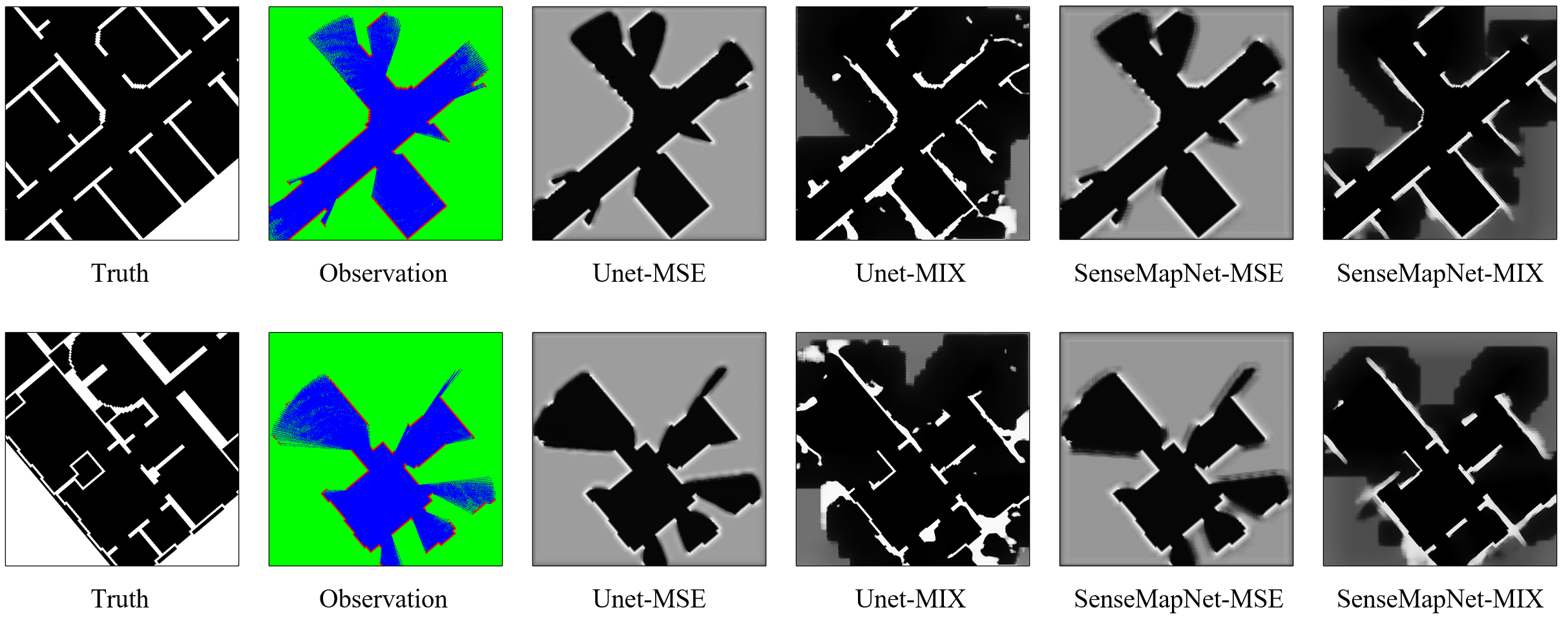}}
	\caption{\textbf{Qualitative comparison of different loss functions and models for local map prediction.}
		The first two columns display the ground truth label map and the corresponding observation map. The remaining columns show the predicted maps generated by different models using either mean squared error (MSE) loss or a hybrid loss (MIX), which combines perceptual and MSE losses. As observed, models trained solely with MSE loss tend to be overly conservative due to the class imbalance between free space and obstacles in the training data. The hybrid loss helps mitigate this issue, improving structure preservation and enhancing the quality of the predicted maps.}
	\label{fig4}
\end{figure}

\begin{figure}[htbp]
	\centerline{\includegraphics[width=0.5\textwidth]{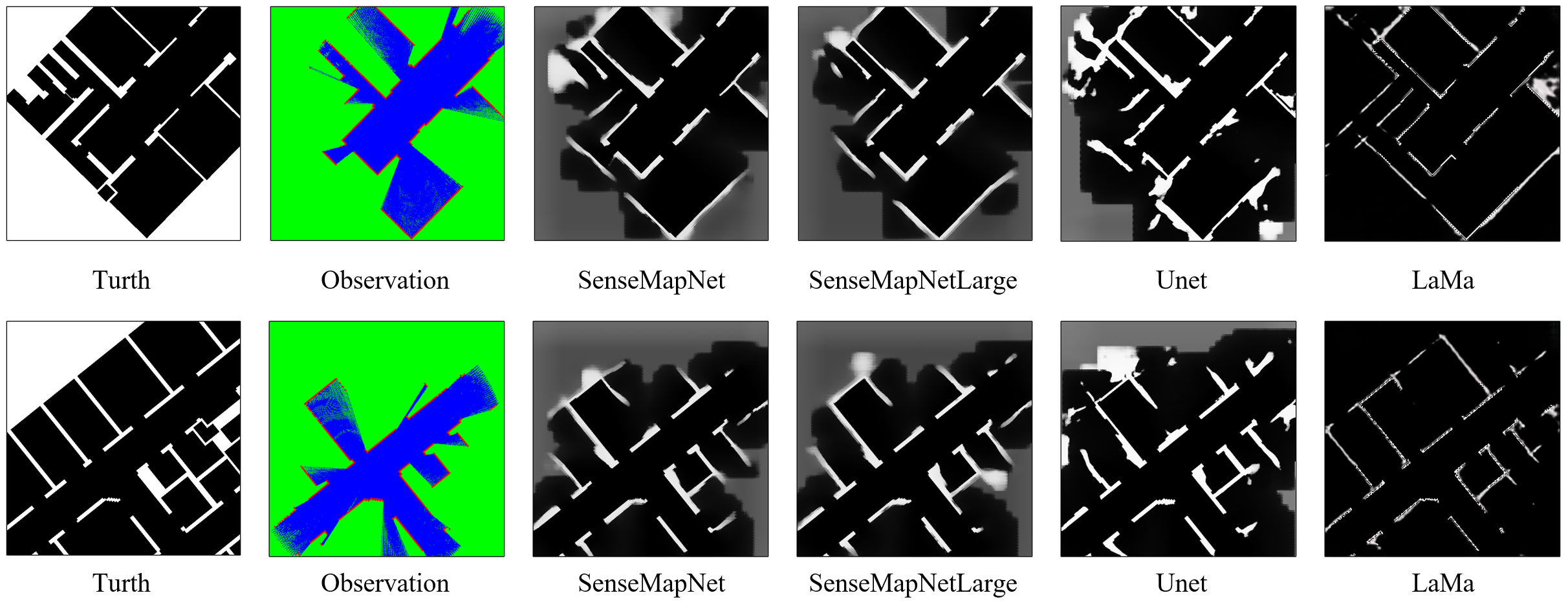}}
	\caption{\textbf{Qualitative comparison of different models on local map prediction.}
		The first two columns represent the ground truth map and the corresponding observation map. The remaining columns display the predicted maps generated by various models, including \emph{SenseMapNet}, \emph{SenseMapNetLarge}, UNet, and LaMa-Fourier. These visual results demonstrate the effectiveness of different architectures in predicting local maps. Notably, \emph{SenseMapNet} and its larger variant show superior structure preservation and spatial consistency compared to UNet and LaMa-Fourier. A detailed quantitative comparison of these models, including the number of parameters, SSIM, LPIPS, and FID, is presented in Tab.~\ref{tab5}.}
	\label{fig5}
\end{figure}
The class imbalance in training data -  where free areas are significantly larger than obstacles - leads to overly conservative predictions when using only mean squared error (MSE) loss, as shown in Fig.~\ref{fig4}.
To mitigate this geometric distortion while maintaining reconstruction fidelity, we implemented a dual-branch architecture that synergistically combines perceptual loss with traditional MSE. This hybrid approach addresses both pixel-level accuracy and structural coherence through its complementary loss components. Performance was evaluated using three metrics: SSIM~\cite{wang2004image}, LPIPS~\cite{zhang2018unreasonable}, and FID ~\cite{bynagari2019gans}.

SSIM evaluates image similarity by considering three key components: luminance, contrast, and structural information. It computes the local statistical features of these three components to quantify the similarity between two images.

The formulation is provided in Eq.~\ref{eq8},
where $\mu_{\hat{m}}$ and $\mu_{m}$ denote the mean luminance of predicted map $\hat{m}$ and ground truth $m$, respectively. The terms $\sigma^2_{\hat{m}}$ and $\sigma^2_{m}$ represent local contrast through their variances, while $\sigma_{\hat{m}m}$ quantifies structural correlation via covariance. Constants $C_1$ and $C_2$ are stabilization parameters preventing division by near-zero values, typically set as $(k_1 L)^2$ and $(k_2 L)^2$ where $L$ is the dynamic range of pixel values. This formulation balances luminance consistency, contrast preservation, and structural alignment—three perceptual dimensions critical for evaluating spatial prediction quality in navigation-oriented applications.

LPIPS, based on the AlexNet model \cite{krizhevsky2012imagenet}, is a metric for evaluating perceptual differences between images, as expressed in Eq.~\ref{eq9}. FID is a widely used metric for evaluating the similarity between generated and real images. It leverages the Inception model~\cite{szegedy2016rethinking} to extract deep image features, computing their mean values $\mu_x$, $\mu_y$ and covariance matrices $\Sigma_x$, $\Sigma_y$. The Fréchet distance is then used to quantify the difference between these two feature distributions, providing a measure of how closely the generated images resemble the real ones, as formulated in Eq.~\ref{eq10}. To ensure accurate evaluation, we utilize the implementation provided in~\cite{seitzer2020pytorch} for FID computation.

\begin{equation} \label{eq8}
	SSIM(\hat{m}, m) =
	\frac{(2\delta_{\hat{m}} \delta_{m} + C_1)(2\sigma_{\hat{m}m} + C_2)}
	{(\delta_{\hat{m}}^2 + \delta_{m}^2 + C_1)(\sigma_{\hat{m}}^2 + \sigma_{m}^2 + C_2)}
\end{equation}

\begin{equation} \label{eq9}
	{LPIPS}(\hat{m}, m) = \sum_j \| \varphi_j(\hat{m}) - \varphi_j(m) \|_2
\end{equation}

\begin{equation} \label{eq10}
	{FID}(x, y) = \| \mu_x - \mu_y \|^2 + \text{Tr}\left( \Sigma_x + \Sigma_y - 2(\Sigma_x \Sigma_y)^{1/2} \right)
\end{equation}

The comparative experimental results are presented in Tab.~\ref{tab4}, while the qualitative results are illustrated in Fig.~\ref{fig4}. Our proposed \emph{SenseMapNet} model demonstrates significant improvements across the SSIM, LPIPS, and FID metrics. In particular, the incorporation of hybrid loss leads to a noticeable enhancement in the SSIM and LPIPS scores for the UNet model as well, indicating the effectiveness of hybrid loss training for this task.

\begin{table}[htbp]
	\caption{Comparison of MSE and Hybrid Loss} \label{tab4}
	\begin{center}
		\resizebox{\columnwidth}{!}{
			\begin{tabular}{|c|c|c|c|c|c|c|}
				\hline
				\textbf{Model}     & \multicolumn{3}{|c|}{\textbf{MSE}} & \multicolumn{3}{|c|}{\textbf{Hybrid Loss}}                                                                                                                  \\
				\cline{2-7}
				                   & \textbf{SSIM} $\uparrow$           & \textbf{LPIPS} $\downarrow$                & \textbf{FID} $\downarrow$ & \textbf{SSIM} $\uparrow$ & \textbf{LPIPS} $\downarrow$ & \textbf{FID} $\downarrow$ \\
				\hline
				Unet               & 0.36                               & 0.82                                       & 276.46                    & 0.72                     & 0.74                        & 281.99                    \\
				\hline
				\emph{SenseMapNet} & 0.36                               & 0.83                                       & 271.31                    & 0.78                     & 0.68                        & 239.79                    \\
				\hline
			\end{tabular}
		}
	\end{center}
\end{table}

\subsection{Comparison of model performance} \label{sec 5.3}
\begin{figure*}[htbp]
	\centerline{\includegraphics[width=1\textwidth]{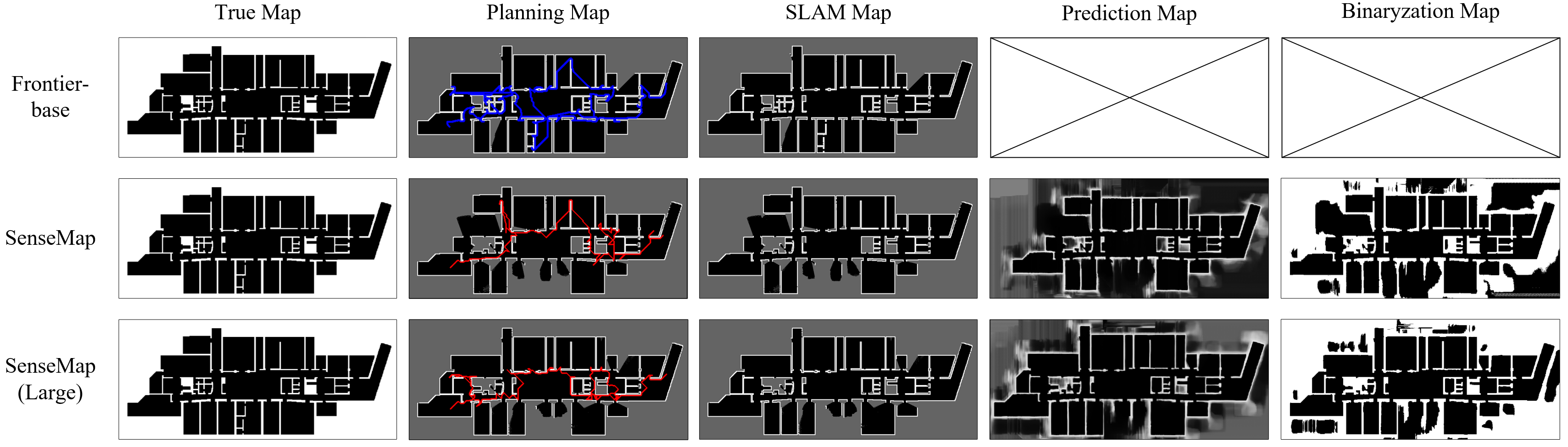}}
	\caption{\textbf{Comparison of exploration methods in map reconstruction.}
		The figure presents the reconstruction results of different exploration strategies. The first column shows the ground truth real maps. The second and third columns depict the planning and SLAM maps, respectively, with the explored areas marked in blue. The fourth column illustrates the predicted maps generated by different methods, where the Frontier-base method does not produce predictions. The final column displays the binarized maps obtained from the predicted outputs. The \emph{SenseMapNet} and \emph{SenseMapNetLarge} models generate significantly more complete and accurate predictions compared to the traditional Frontier-base approach, demonstrating their effectiveness in autonomous exploration.}
	\label{fig6}
\end{figure*}
In order to compare the performance of the models at different scales, we conducted experiments using the hybrid loss of Mean Squared Error (MSE) and Perceptual Loss. We compared the number of parameters, SSIM, LPIPS, and FID between UNet, \emph{SenseMapNet} and \emph{SenseMapNetLarge}. We also trained LaMa-Fourier \cite{suvorov2022resolution} on the \emph{SenseMapDataset} using its original training protocol as a baseline comparison. Results are shown in Tab.~\ref{tab5} with visual comparisons in Fig.~\ref{fig5}.

\begin{table}[htbp]
	\caption{Comparison of Different Models} \label{tab5}
	\begin{center}
		\begin{tabular}{|c|c|c|c|c|}
			\hline
			\textbf{Model}          & \textbf{Params. (M)} & \textbf{SSIM} $\uparrow$ & \textbf{LPIPS} $\downarrow$ & \textbf{FID} $\downarrow$ \\
			\hline
			LaMa-Fourier            & 27.04                & 0.67                     & 0.40                        & 240.268                   \\
			\hline
			Unet                    & 43.57                & 0.72                     & 0.69                        & 232.74                    \\
			\hline
			\emph{SenseMapNetLarge} & 39.35                & 0.77                     & 0.67                        & 228.14                    \\
			\hline
			\emph{SenseMapNet}      & 14.36                & 0.78                     & 0.68                        & 239.79                    \\
			\hline
		\end{tabular}
	\end{center}
\end{table}

These results demonstrate that \emph{SenseMapNet} achieves superior performance while maintaining a lightweight model architecture. Compared to UNet, it exhibits improvements across all four key metrics, including parameter efficiency, SSIM, LPIPS, and FID. Furthermore, in comparison with LaMa, which leverages adversarial learning, \emph{SenseMapNet} achieves better results in three critical aspects: parameter efficiency, SSIM, and FID.

\subsection{Comparison of reconstruction efficiency} \label{sec 5.4}
We conducted experiments on 10 different maps, with each map being explored 50 times. Each exploration started with different unknown regions. We compared the performance of our method with different scale models and the traditional Frontier-based method, using three evaluation metrics: exploration time, exploration coverage $\rho$, and reconstruction accuracy. The formulas for exploration coverage $\rho$ and RA (reconstruction accuracy) are given below Eq.~\ref{eq12}, where $I({condition})$ is the indicator function, which equals 1 when the condition is True and 0 when it is False.

\begin{figure}[htbp]
	\centerline{\includegraphics[width=0.5\textwidth]{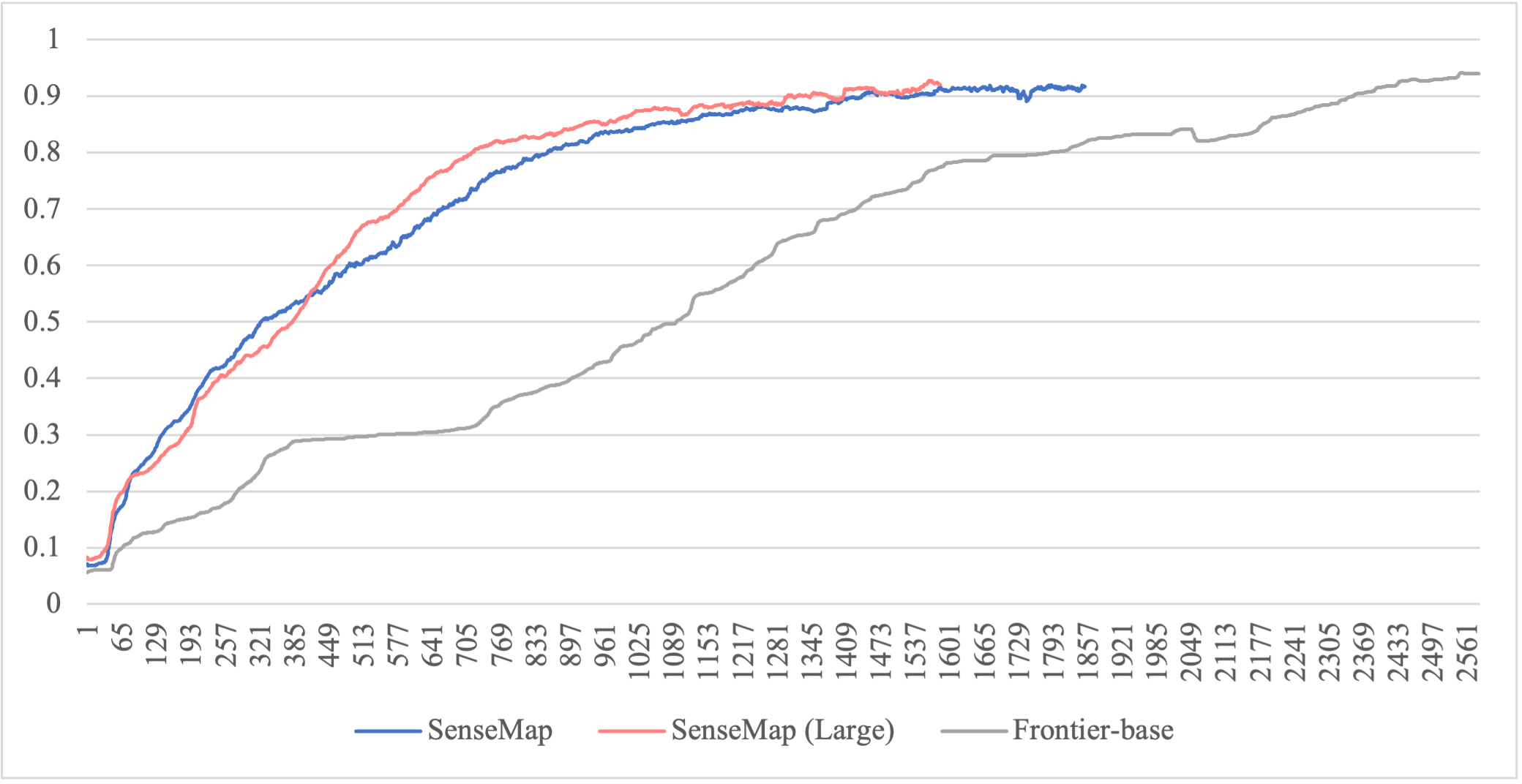}}
	\caption{\textbf{Exploration coverage over time.}
		The graph illustrates the variation of exploration coverage $\rho$ as a function of time for different exploration methods.
		The \emph{SenseMapNet} and \emph{SenseMapNetLarge} models exhibit significantly faster coverage growth compared to the conventional Frontier-base method, indicating improved exploration efficiency.
		Quantitative comparisons of average exploration time, coverage, and reconstruction accuracy are provided in Tab.~\ref{tab6}.}
	\label{fig7}
\end{figure}

\begin{equation} \label{eq11}
	\rho = \frac{\sum_{a,b \in R} I(M(a,b) \text{ is free})}{\sum_{a,b \in R} I(M_{{true}}(a,b) \text{ is free})} \times 100\%
\end{equation}

\begin{equation} \label{eq12}
	\resizebox{\columnwidth}{!}{$
	{RA} = \frac{\sum_{a,b \in R} (I(M(a,b) \text{ is free}) \times I(M_{{true}}(a,b) \text{ is free}))}{\sum_{a,b \in R} I(M(a,b) \text{ is free})} \times 100\%
		$}
\end{equation}

\begin{table}[htbp]
	\caption{Comparison of Different Exploration Methods}
	\label{tab6}
	\centering
	\renewcommand{\arraystretch}{1.2}
	\begin{tabular}{|c|c|c|c|}
		\hline
		\textbf{Method}      & \makecell{\textbf{Avg.}               \\ \textbf{Exploration Time}} & \makecell{\textbf{Avg.} \\ \textbf{Coverage} $\rho$} & \textbf{Avg. RA} \\
		\hline
		Frontier-base        & 2,335.56                & 0.91 & -    \\
		\hline
		\emph{SenseMap}      & 1,248.68                & 0.88 & 0.88 \\
		\hline
		\emph{SenseMapLarge} & 1,166.50                & 0.90 & 0.90 \\
		\hline
	\end{tabular}
\end{table}

The reconstruction results of the four exploration methods are illustrated in Fig.~\ref{fig6}, while the variation of exploration coverage $\rho$ over time is depicted in Fig.~\ref{fig7}. The average exploration time, mean exploration coverage $\rho$, and average reconstruction accuracy across all experiments are summarized in Tab.~\ref{tab6}.

From the experimental results, it is evident that the \emph{SenseMap} method significantly reduces the time required for exploration while improving exploration efficiency.

\section{Conclusion} \label{sec 6}
In this study, we introduced \emph{SenseMap}, a novel approach that leverages neural network-based prediction to enhance the efficiency of autonomous exploration in structured indoor environments. By integrating \emph{SenseMapNet}, a lightweight hybrid architecture combining convolutional encoders with transformer-based perception, our method enables robots to infer unobserved regions and optimize exploration trajectories in real time. Through extensive experiments on \emph{SenseMapDataset}, we demonstrated that our approach outperforms conventional exploration strategies, reducing exploration time while maintaining high reconstruction accuracy.

Comparative evaluations revealed that \emph{SenseMapNet} achieves superior performance across multiple perceptual quality metrics, including SSIM, LPIPS, and FID, while maintaining a lightweight model suitable for onboard computation. Furthermore, our approach outperforms traditional frontier-based methods by effectively leveraging predicted environmental structures to guide exploration.

Future research will focus on extending \emph{SenseMap} to multi-robot systems, enabling collaborative exploration and decentralized decision-making. Additionally, integrating uncertainty-aware learning frameworks and reinforcement learning-based policies could further enhance adaptive exploration in dynamically changing environments. These advancements will pave the way for more intelligent and efficient robotic exploration in real-world applications.

\section*{Acknowledgment}

This work is supported by Key Research and Development Program of Xinjiang Uygur Autonomous Region (No.2022B01008-3), National Natural Science Foundation of China under grant No. 92164203, 62334006, Beijing National Research Center for Information Science and Technology, and Sichuan Science and Technology Program under Grant (2023YFG0302).




\bibliographystyle{IEEEtran}
\bibliography{refs}

\end{document}